\documentclass[10pt,twocolumn,letterpaper]{article}

\usepackage[pagenumbers]{cvpr}
\usepackage[accsupp]{axessibility} 
\usepackage{graphicx, amsmath, amssymb, caption, subcaption, multirow, overpic, textpos}
\usepackage[table]{xcolor}
\usepackage[pagebackref=false, breaklinks=true, letterpaper=true, colorlinks, citecolor=citecolor, linkcolor=linkcolor, bookmarks=false]{hyperref}
\definecolor{citecolor}{HTML}{0071BC}
\definecolor{linkcolor}{HTML}{ED1C24}

\def\cvprPaperID{5135}
\def\confName{CVPR}
\def\confYear{2023}

\makeatletter
\def\thanks#1{\protected@xdef\@thanks{\@thanks
        \protect\footnotetext{#1}}}
\makeatother

\newcommand{\authorskip}{\hspace{5mm}}

\begin{document}

\title{Affordance Grounding from Demonstration Video to Target Image \vspace{-3mm}}

\author{Joya Chen \authorskip Difei Gao \authorskip Kevin Qinghong Lin \authorskip Mike Zheng Shou$^{\dagger}$\thanks{$^{\dagger}$Corresponding Author.}\\[1mm]
Show Lab, National University of Singapore \\
\footnotesize \texttt{\{joyachen,qinghonglin\}@u.nus.edu \authorskip \{daniel.difei.gao,mike.zheng.shou\}@gmail.com}
}

\maketitle

\begin{abstract}
Humans excel at learning from expert demonstrations and solving their own problems. To equip intelligent robots and assistants, such as AR glasses, with this ability, it is essential to ground human hand interactions (\ie, affordances) from demonstration videos and apply them to a target image like a user's AR glass view. This video-to-image affordance grounding task is challenging due to (1) the need to predict fine-grained affordances, and (2) the limited training data, which inadequately covers video-image discrepancies and negatively impacts grounding. To tackle them, we propose Affordance Transformer (Afformer), which has a fine-grained transformer-based decoder that gradually refines affordance grounding. Moreover, we introduce Mask Affordance Hand (MaskAHand), a self-supervised pre-training technique for synthesizing video-image data and simulating context changes, enhancing affordance grounding across video-image discrepancies. Afformer with MaskAHand pre-training achieves state-of-the-art performance on multiple benchmarks, including a substantial 37\% improvement on the OPRA dataset. Code is made available at \hyperlink{https://github.com/showlab/afformer}{https://github.com/showlab/afformer}.
\end{abstract}

\section{Introduction}

\begin{figure}[t]
    \centering
    \includegraphics[width=\linewidth]{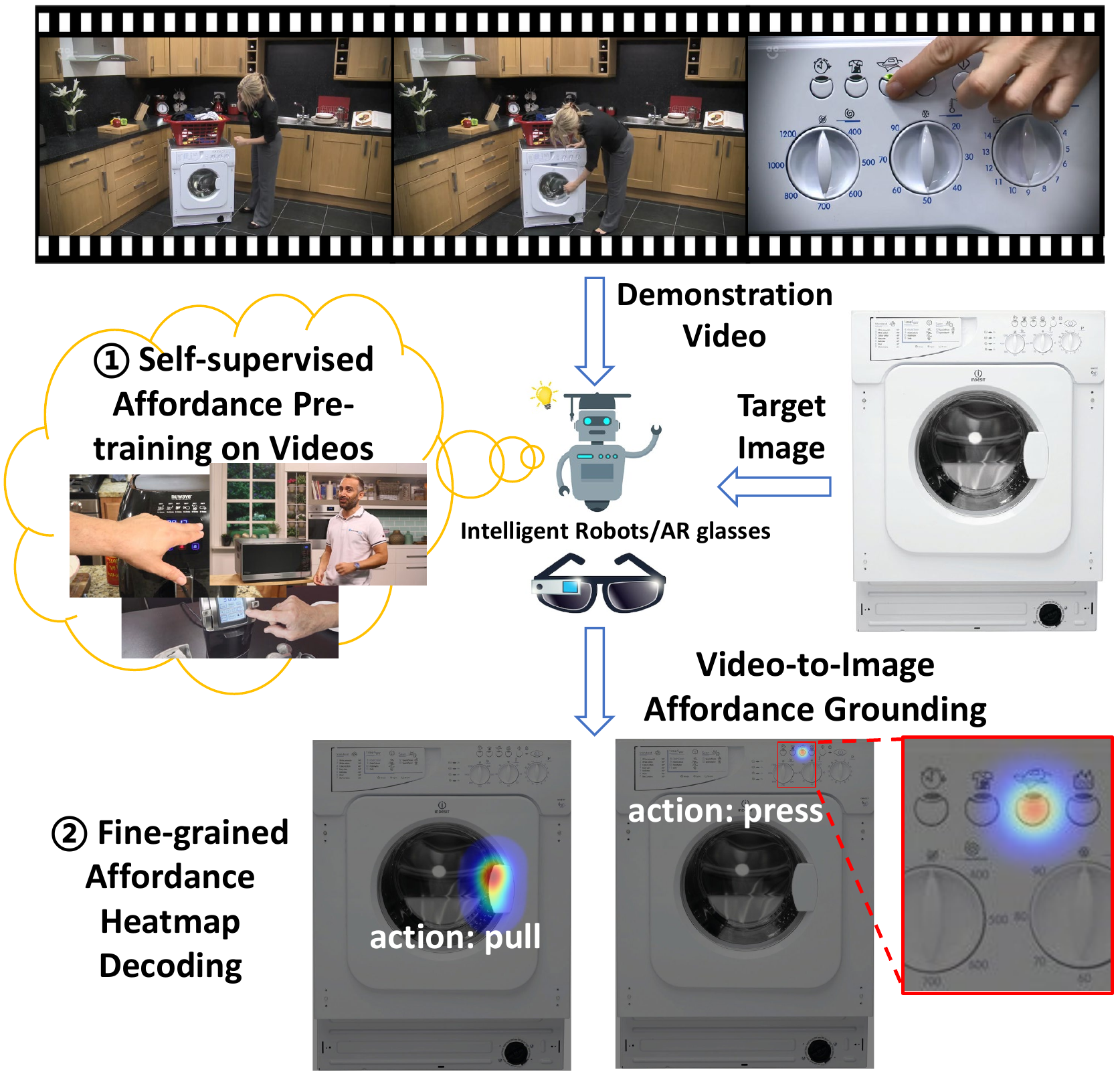}
    \vspace{-0.6cm}
    \caption{This figure demonstrates the video-to-image affordance grounding task, which aims to identify the area of human hand interaction (\ie, affordance) in a demonstration video and map it to a target image (\eg, AR glass view). Our contributions include (1) proposing a self-supervised pre-training approach for affordance grounding, and (2) establishing a new model that excels remarkably in fine-grained heatmap decoding. \vspace{-0.6cm}}
    \label{figure1}
\end{figure}

Humans frequently learn from observing others interact with objects to enhance their own experiences, such as following an online tutorial to operate a novel appliance. To equip AI systems with this ability, a key challenge lies in comprehending human interaction across videos and images. Specifically, a robot must ascertain the points of interaction (\ie, affordances~\cite{affo_survey}) in a demonstration video and apply them to a new target image, such as the user's view through AR glasses.

This process is formulated as video-to-image affordance grounding, recently proposed by \cite{demo2vec}, which presents a more challenging setting than previous affordance-related tasks, including affordance detection\cite{affonet,oneshot_affo}, action-to-image grounding~\cite{hotspot,hagnet,probes,exoaffo}, and forecasting~\cite{jointhand,nextactive,forecast_hoi}. The complexity of this setting stems from two factors: (1) Fine-grained grounding: Unlike conventional grounding tasks that usually localize coarse affordance positions (\eg, identifying all buttons related to ``press''), video-to-image affordance grounding predicts fine-grained positions specific to the query video (\eg, only buttons pressed in the video). (2) Grounding 
accross various video-image discrepancies: Demonstration videos and images are often  captured in distinct environments, such as a store camera's perspective versus a user's view in a kitchen, which complicates the grounding of affordances from videos to images. Moreover, annotating for this task is labor-intensive, as it necessitates thoroughly reviewing the entire video, correlating it with the image, and pinpointing affordances. As a result, affordance grounding performance may be limited by insufficient data on diverse video-image discrepancies.

To enable fine-grained affordance grounding, we propose a simple yet effective Affordance transformer (Afformer), which progressively refines coarse-grained predictions into fine-grained affordance grounding outcomes. Previous methods~\cite{demo2vec,hagnet,hotspot} either simply employ large stride upsampling or deconvolution for coarse-grained affordance heatmap prediction (\eg, $8\times8 \rightarrow 256\times256$~\cite{demo2vec}), or just evaluate at low resolution (\eg, $28\times28$~\cite{hotspot,hagnet}). As a result, these methods struggle with fine-grained affordance grounding, particularly when potential affordance regions are closely situated (\eg, densely packed buttons on a microwave). Our approach employs cross-attention~\cite{detr,mask2former} between multi-scale feature pyramids, to facilitate gradual decoding of fine-grained affordance heatmaps.

To address the limited data issue that inadequately covers video-image differences and hampers affordance grounding performance, we present a self-supervised pre-training method, Masked Affordance Hand (MaskAHand), which can leverage vast online videos to improve video-to-image affordance grounding. MaskAHand can automatically generate target images from demonstration videos by masking hand interactions and simulating contextual differences between videos and images. In the generated target image, the task involves estimating the interacting hand regions by watching the original video, thereby enhancing the similarity capabilities crucial for video-to-image affordance grounding. Our approach uniquely simulates context changes, an aspect overlooked by previous affordance pre-training techniques~\cite{jointhand,probes}. Furthermore, we also rely less on external off-the-shelf tools~\cite{100doh,sift,cvzone,skinseg} used in \cite{jointhand,probes}.

We conducted comprehensive experiments to evaluate our Afformer and MaskAHand methods on three video-to-image affordance benchmarks, namely OPRA~\cite{demo2vec}, EPIC-Hotspot~\cite{hotspot}, and AssistQ~\cite{assistq}. Compared to prior architectures, our most lightweight Afformer variant achieves a relative $33\%$ improvement in fine-grained affordance grounding ($256\times256$) on OPRA and relative gains of 10\% to 20\% for coarse-grained affordance prediction ($28\times28$) on OPRA and EPIC-Hotspot. Utilizing MaskAHand pre-training for Afformer results in zero-shot prediction performance that is comparable to previously reported fully-supervised methods on OPRA~\cite{demo2vec}, with fine-tuning further enhancing the improvement to $37\%$ relative gains. Moreover, we demonstrate the advantage of MaskAHand when the data scale for downstream tasks is limited: on AssistQ, with only approximately 600 data samples, MaskAHand boosts Afformer's performance by a relative $28\%$ points.

\section{Related Work}

\noindent\textbf{Visual Affordance.} The term ``affordance'' was coined in 1979~\cite{affordance1979} to refer to potential action possibilities offered by the environment. With the progress of computer vision, learning visual affordances has gained significant attention in several domains, such as 2D/3D human-object interaction~\cite{affonet,3daffonet,graph_affo,100doh}, robotic manipulation/grasping~\cite{affo4robot_survey,dexaffo,dexvip}, and instructional video understanding~\cite{demo2vec,hotspot}. Affordance detection~\cite{oneshot_affo,affonet}, forecasting~\cite{jointhand,nextactive,forecast_hoi}, and grounding~\cite{hotspot,hagnet,probes,exoaffo,demo2vec} are some of the well-known visual affordance tasks.

\noindent\textbf{Affordance Grounding.} Affordance grounding tasks are primarily divided into two categories: (1) Action-to-Image Grounding~\cite{hotspot,hagnet,probes,exoaffo}, which aims to identify image regions corresponding to a specific action query (\eg, ``press'' $\rightarrow$ all microwave buttons). (2) Video-to-Image Affordance Grounding~\cite{demo2vec,hotspot,hagnet}, which involves predicting the interaction region and associated action label in a target image, based on a video (\eg, pressing the power button on a microwave in a video  $\rightarrow$ the same region in a target image). Our paper concentrates on the latter problem, which presents greater challenges than action-to-image grounding.

\noindent\textbf{Predicting Affordance Heatmap.}
Affordance regions have typically been represented by pixel-wise heatmaps or masks in prior research~\cite{demo2vec,hotspot,hagnet,probes,exoaffo,oneshot_affo,affonet}. To predict these pixel-wise affordance maps, upsampling and deconvolution with a large stride (\eg, $32\times$) are frequently employed~\cite{demo2vec,probes,oneshot_affo}. Additionally, some weakly-supervised approaches~\cite{hotspot,hagnet,exoaffo} utilize Grad-CAM~\cite{gradcam} to estimate the affordance heatmap through gradient activations. However, these techniques struggle to generate fine-grained affordance heatmaps, as they only learn affordance in low-resolution feature maps (\eg, $7\times7$).

We get inspiration from multi-scale segmentation transformers~\cite{maskformer,mask2former,pvt,segformer} to design a pyramid transformer decoder, which performs cross attention across pyramid feature levels to gradually refine the affordance heatmap estimations. This architecture can better support fine-grained affordance heatmap prediction.

\noindent\textbf{Self-supervised Affordance Pre-training.} Affordance regions within or across objects can be highly diverse and irregular in shape, complicating the prediction process. Several affordance pre-training methods~\cite{jointhand,probes} have been introduced to improve affordance heatmap prediction. HoI-forecast~\cite{jointhand} generates self-supervised annotations for the current frame using future frames by: (1) employing hand-object interaction detection~\cite{100doh}, skin segmentation~\cite{skinseg}, and hand landmark estimation~\cite{cvzone} to produce affordance points, and (2) utilizing SIFT~\cite{sift} matching to map affordance points back to the current frame. HandProbes~\cite{probes} also leverages \cite{100doh} to select frames for ``masked hand-object intersection prediction''.

In contrast to HoI-forecast~\cite{jointhand} and HandProbes~\cite{probes}, which primarily focus on minimally altered egocentric images/videos~\cite{egovlp} and rely on detecting common objects, our proposed MaskAHand enhances affordance grounding from video to image across substantial contextual variations, eliminating the need for common object detection.

\begin{figure*}[t]
    \centering
    \includegraphics[width=\linewidth]{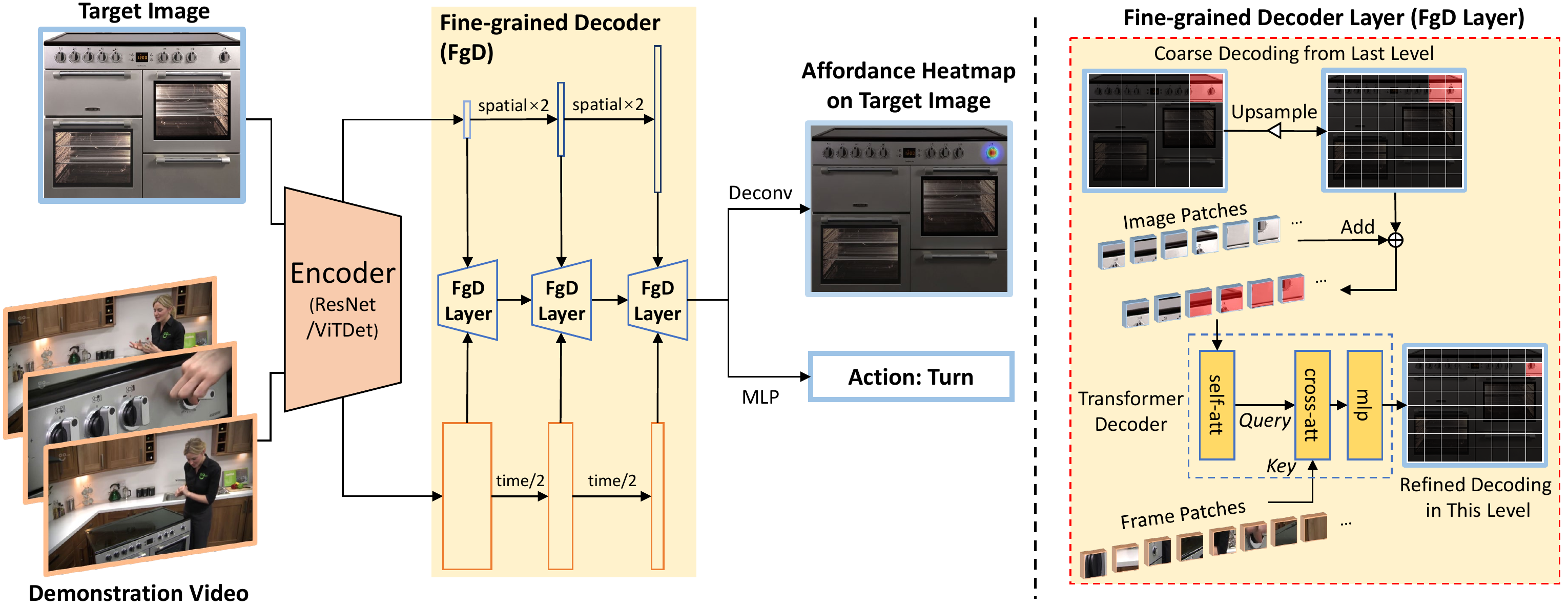}
    \caption{Our proposed Afformer is a simple yet effective model for video-to-image affordance grounding. The Afformer takes a demonstration video and a target image as inputs and produces an affordance heatmap on the target image. It employs an encoder to extract video and image features, followed by a multi-scale, transformer-based decoder to progressively refine fine-grained affordance heatmaps.}
    \label{figure2}
\end{figure*}

\section{Affordance Transformer}

We introduce Affordance Transformer (Afformer), a novel model for video-to-image affordance grounding. See Figure~\ref{figure2}, our model employs a multi-scale, transformer-based decoder to handle fine-grained affordance grounding. We now proceed to discuss the Afformer in detail.

\subsection{Formulation} \label{section3.1}

We present the problem formulation for video-to-image affordance grounding~\cite{demo2vec}. Given a demonstration video $V \in \mathbb{R}^{t\times 3 \times h^V \times w^V}$ and a target image $I \in \mathbb{R}^{3 \times h^I \times w^I}$ as inputs, the goal of video-to-image affordance grounding is to predict an affordance heatmap $H \in \mathbb{R}^{h^I \times w^I}$ over the target image, accompanied by an action label $A \in \{1, \ldots, c\}$, where $c$ denotes the number of action classes. Here, $t$ represents the number of sampled video frames, $h$ and $w$ indicate the height and width of the image or video, respectively. The Afformer $\boldsymbol{F}$ can be described as a function that maps the demonstration video and target image to the affordance heatmap and action label:

\begin{equation}
\boldsymbol{F}(V, I) \rightarrow (H, A). \label{equation1}
\end{equation}

\subsection{Afformer Architecture}

As shown in Figure~\ref{figure2}, the Afformer $\boldsymbol{F}$ consists of a video/image shared encoder $\boldsymbol{F_e}$, a fine-grained decoder $\boldsymbol{F_d}$, and a prediction head $\boldsymbol{F_h}$, \ie $\boldsymbol{F} = \boldsymbol{F_h} \circ (\boldsymbol{F_d} \circ \boldsymbol{F_e})$. 

\noindent\textbf{Shared Encoder.} Our Afformer employs a shared encoder for both video and image feature representation, mitigating overfitting issues in small video-to-image affordance grounding datasets~\cite{demo2vec,hotspot,assistq}. During the encoding phase, we only utilize a spatial network~\cite{resnet,vit} to process images and sampled video frames, reducing computational costs caused by temporal modeling. To enable multi-scale decoding, we preserve multi-scale features during the encoding process. This can be expressed as:

\begin{equation}
\boldsymbol{F_e}(V) \rightarrow \{E_l^V\}, \boldsymbol{F_e}(I) \rightarrow \{E_l^I\}, \label{equation2}
\end{equation}

\noindent where $\{E_l^V\}$ ($\{E_l^I\}$) represents the set of features extracted from the video (image) input at different scales $l$. Their shapes are  $E^l_V \in \mathbb{R}^{t\times C\times h_l^V \times w_l^V}$ and $E^l_I \in \mathbb{R}^{C\times h_l^I \times w_l^I}$. $C$ is the feature dimension. Following \cite{fpn}, $l$ is the stride with respect to the input scale, \eg $l \in \{2,3,4,5\}$. We adopt multi-scale feature pyramid networks such as ResNet with FPN~\cite{resnet,fpn} and ViTDet~\cite{vit,vitdet} as the encoder.

\noindent\textbf{Fine-grained Affordance Decoder.}
With multi-scale encodings $\{E_l^V\}$ and $\{E_l^I\}$, our fine-grained decoder produces heatmap decoding $D_{l_{min}}$ by

\begin{equation}
    \boldsymbol{F_d}(\{E_l^V\}, \{E_l^I\}) \rightarrow D_{l_{min}}. \label{equation3}
\end{equation}

\noindent where $l_{min}$ represents the minimal stride, which corresponds to the pyramid level of the highest feature resolution. For example, if the encoder uses $l_{min} = 2$, then $D_{l_{min}}$ will be in $\mathbb{R}^{C \times (h^I / 2^2) \times (w^I / 2^2)}$. Previous methods~\cite{demo2vec, hotspot} typically use single-step, small-scale decoding (\eg, $l_{min} = 5$) and a large deconvolution to predict large-scale heatmaps. However, this approach loses spatial information and produces coarse predictions. In comparison, our decoder utilizes a multi-scale strategy to gradually decode the heatmap, which leads to more detailed affordance heatmaps.

We first outline single-scale decoding before extending it to multi-scale. Video-to-image affordance grounding can be considered a process where, at each spatial location in the target image, we search the video to determine if the location corresponds to an affordance region. Consequently, this process can be intuitively modeled as a cross-attention operation, with image encodings serving as the query and video encodings as the key and value. Unlike previous studies~\cite{demo2vec,hotspot}, our method explicitly incorporates spatial modeling, making it more suitable for heatmap grounding. The core operation of multi-head cross-attention ($MCA$) is

\begin{equation}
\begin{aligned}
\boldsymbol{MCA}(\hat{Q_l},\hat{K_l},\hat{V_l}) = \sigma(\frac{\hat{Q_l}(\hat{K_l})^\mathrm{T}}{\sqrt{C}} + R_l)\hat{V_l}W_l^{MCA},
\end{aligned}
\end{equation}

\noindent where $\hat{Q_l},\hat{K_l},\hat{V_l}$ denote flattened, layer-normalized, and linearly mapped query, key, and value features from $E^I_l,E^V_l,E^V_l$, respectively. $R_l$ represents the decomposed relative positional encoding~\cite{vitdet,mvitv2}. $W^{MCA}$ is a learnable linear mapping, and $\sigma$ is the softmax operation along the sequence of the $\hat{K_l}$ axis. For simplicity, we only display a single attention head here, while the actual attention head used is a common $C / 64$. According to \cite{transformer}, the complete transformer decoder layer also consists of a multi-head self-attention layer ($MSA$) and an MLP layer ($MLP$). We perform $MSA$ on image features, rather than video features, to reduce memory usage, as video feature sequences are longer due to the temporal dimension. The single-scale decoding $D_l$ is obtained by

\begin{equation}
D_l = flatten(E^I_l) + \boldsymbol{MSA}(\hat{Q_l}), \label{equation5} 
\end{equation}
\vspace{-5mm}
\begin{equation}
D_l = D_l + \boldsymbol{MCA}(D_l,\hat{K_l},\hat{V_l}), \label{equation6}
\end{equation}
\vspace{-5mm}
\begin{equation}
D_l = D_l + \boldsymbol{MLP}(D_l). \label{equation7}
\end{equation}

We proceed with multi-scale decoding. The query in Equation~\ref{equation5} should incorporate heatmap decoding at the pyramid level $(l+1)$ for coarse-to-fine refinement. Furthermore, we find that maintaining a fixed resolution video pyramids stabilize training. We select $l=3$ to balance resolution and semantics for video pyramids. Meanwhile, as the encoder only processes videos spatially, we apply temporal sampling to video features to aggregate temporal information before decoding, as depicted in Figure~\ref{figure2}. Consequently, the multi-scale decoding can be expressed as

\begin{equation}
D_l =  flatten(E^I_l) + \boldsymbol{MSA}(\hat{Q_l} + UP(D_{l+1})), \label{equation8} 
\end{equation}
\vspace{-5mm}
\begin{equation}
D_l = D_l + \boldsymbol{MCA}(D_l,\hat{K_l}^{C3D},\hat{V_l}^{C3D}), \label{equation9}
\end{equation}
\vspace{-5mm}
\begin{equation}
D_l = D_l + \boldsymbol{MLP}(D_l). \label{equation10}
\end{equation}

\noindent where $UP$ denotes to the nearest spatial upsampling used in FPN~\cite{fpn}. ${C3D}$ denotes spatial-temporal convolution~\cite{c3d} with a kernel size of 3 and a stride of 2 applied only in the temporal dimension. By multi-scale decoding we can get the final fine-grained heatmap decoding $D_{l_{min}}$.

\noindent\textbf{Heatmap and Action Prediction.} The predictor computes $\boldsymbol{F_h}(D_{l_{\min}}) \rightarrow H, A$. We utilize appropriate deconvolutional layers to reconstruct the heatmap $H$ by decoding $D_{l_{\min}}$ (\eg, $C\times64\times64 \mapsto 1\times256\times256$). A MLP network and adaptive 2D pooling are employed for action classification $A$. Thus, we have concisely presented the Afformer as $\boldsymbol{F}(V, I) \rightarrow H, A$.

\noindent\textbf{Training Loss Function.} Our Afformer predicts action logits for $c$ classes and the corresponding $h^I \times w^I$ heatmap. In a supervised setting, Afformer is trained with action classification loss $\mathcal{L}_a$ and heatmap regression loss $\mathcal{L}_h$. The former uses multi-class cross-entropy loss, while the latter employs KL divergence as the heatmap loss~\cite{demo2vec,saliency_loss}:

\begin{equation}
    \mathcal{L}_{h} = \sum_i^{h^I}\sum_j^{w^I} 1_{g_{i,j}>0} \sigma_{g}(g)_{i,j}\log\frac{\sigma_g(g)_{i,j}}{\sigma_{h}(h)_{i,j}},  \label{equation11}
\end{equation}

\noindent where $g$ represents the ground-truth heatmap, $\sigma_g(g)_{i,j} = g_{i,j} / \sum g$ normalizes by sum, and $\sigma_h(h)_{i,j} = \exp(h_{i,j}) / \sum \exp(h)$ normalizes by softmax. The total loss for optimization is $\mathcal{L} = \mathcal{L}_{a} + \mathcal{L}_{h}$.

Our proposed Afformer attains fine-grained heatmap decoding via multi-scale cross-attention, making it an effective model for video-to-image affordance grounding. However, the limited available training data~\cite{demo2vec,hotspot,assistq} ($<20k$) restricts context variation and may impair the Afformer's ability to ground affordances across video-image discrepancies. While some modified cross-attention techniques~\cite{crossattn_fewshot,cat} aim to address the low-data issue~\cite{crossattn_fewshot}, they focus on feature aspects and are limited to few-shot settings. We propose to tackle the problem from data, yielding a more versatile solution applicable to various challenges. In the following section, we introduce a self-supervised pre-training method for video-to-image affordance grounding.

\begin{figure*}[t]
    \centering
    \includegraphics[width=\linewidth]{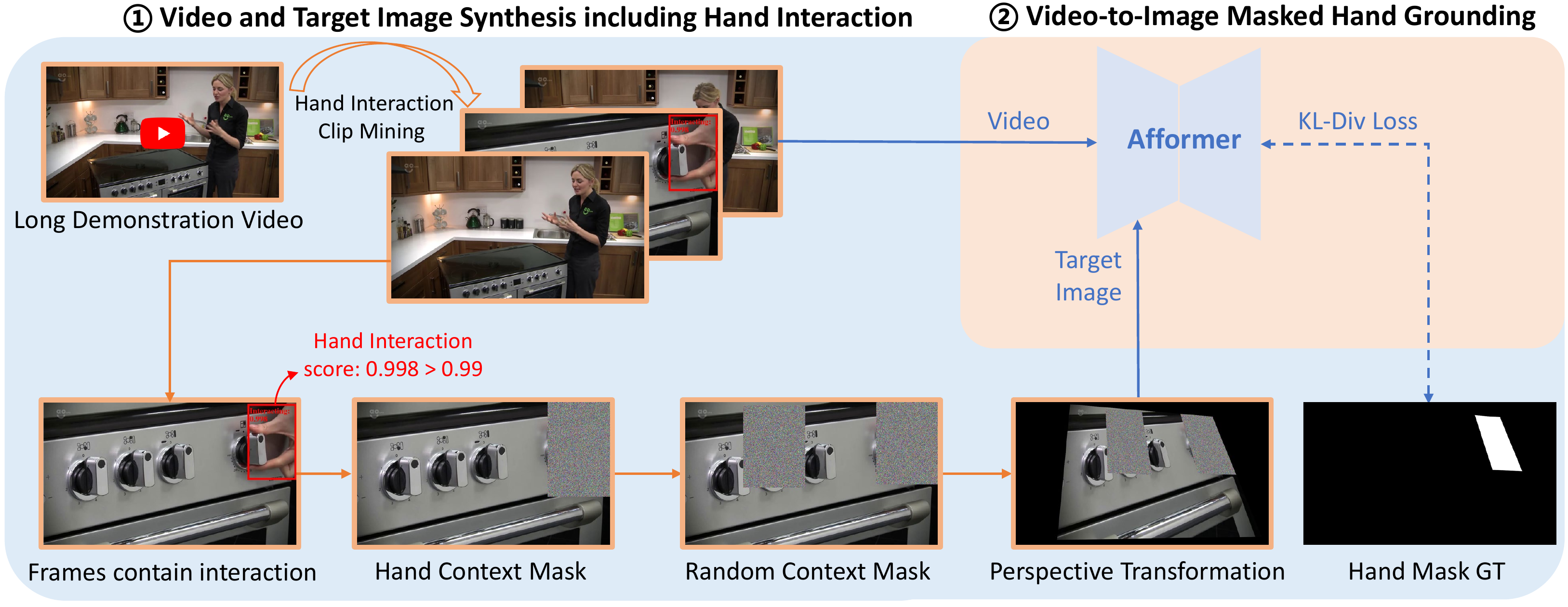}
    \caption{MaskAHand pre-training focuses on ``video-to-image masked interaction hand grounding'', acting as a proxy task for video-to-image affordance grounding. There are two steps: (1) video and target image synthesis including hand interaction, and (2) training Afformer with the generated data to learn ``video-to-image masked interaction hand grounding''. GT stands for ``ground-truth''.}
    \label{figure3}
\end{figure*}

\section{Masked Affordance Hand Grounding}

We introduce Masked Affordance Hand (MaskAHand), a self-supervised pre-training approach for video-to-image affordance grounding. As depicted in Figure~\ref{figure3}, our method leverages the consistency between hand interaction and affordance regions, approximating affordance grounding to hand interaction grounding. MaskAHand can be viewed as an extension of the ``masked image modeling'' paradigm~\cite{maskfeat,mae,oatrans} to ``masked hand grounding''. We now elaborate on the MaskAHand method in detail.

\subsection{Formulation} \label{section4.1}

As described in Section~\ref{section3.1}, the video-to-image affordance grounding can be expressed as Equation~\ref{equation1}: $\mathbf{F}(V, I) \rightarrow H, A$. 
Since action prediction is a well-established problem in action recognition~\cite{kinetics400,c3d,slowfast,stmae,morphmlp}, we focus on the challenging task of affordance heatmap grounding, denoted by $\boldsymbol{F}(V, I) \rightarrow H$. However, annotating for this heatmap is labor-intensive, as it necessitates thoroughly reviewing the entire video, correlating it with the image, and pinpointing affordances. Consequently, current video-to-image affordance grounding datasets contain limited training samples, with fewer than $20k$ in total.

We consider solving the limited data problem in the task $\boldsymbol{F}(V, I) \rightarrow H$. We observe that all affordance regions are interacted with by hands. The human hand exhibits a distinct visual pattern, making it easier to detect than irregular affordance regions. The interaction state can also be readily distinguished using the hand interaction detector~\cite{100doh}. Consequently, we focus on a related task, $\boldsymbol{F'}(V,I) \rightarrow H'$, which is simpler to gather data for, where $H'$ represents the "imagined" hand in the target image that interacts with the affordance region shown in the video. As demonstrated, the capabilities of $\boldsymbol{F'}$ and $\boldsymbol{F}$ are closely related, allowing us to obtain $\boldsymbol{F}$ by fine-tuning the $\boldsymbol{F'}$ network or even considering $\boldsymbol{F'}$ as $\boldsymbol{F}$ for zero-shot grounding.

However, training $\boldsymbol{F'}(V,I)\rightarrow H'$ still needs the demonstration video $V$, the target image $I$, and interaction hand annotation heatmap $H'$ (\eg, hand box mask). But thanks to our approximation for the original task, the data preparation becomes much simpler and allows us to do self-supervised pre-training. We refer to our approach as Masked Affordance Hand (MaskAHand), illustrated in Figure~\ref{figure3} and described in the following section.

\subsection{Affordance-related Data Synthesis}

\noindent\textbf{Hand Interaction Detection.} 
Our MaskAHand relies solely on a hand interaction detector, eliminating the need for an object detector as required by \cite{jointhand,probes}. We employ a Faster R-CNN~\cite{faster_rcnn,fpn} trained on the 100DOH dataset~\cite{100doh,epic-kitchens-55,charades-ego,egtea} to detect hand bounding boxes and output binary hand states (\ie, interacting or not). Our trained detector achieves an 84.9 AP in hand interaction detection on the 100DOH test set, demonstrating its accuracy and reliability for synthesizing hand interaction data.

\noindent\textbf{Hand Interaction Clip Mining.}
We extract multiple hand interaction clips from a long demonstration video, each containing 32 consecutive frames and regarding as $V$, guaranteeing the presence of an interacting hand in at least one frame. To avoid redundancy, we apply a stride of 16 frames between successive clips. We only set a high interaction score threshold 0.99 to reduce false positives.

\noindent\textbf{Target Image Synthesis and Transformation.} 
Inspired by SuperPoint~\cite{superpoint}, we synthesize the target image $I$ from $V$ by simulating video-image context changes, as illustrated in Figure~\ref{figure3}. The process consists of four steps: (1) Select the corresponding frame from $V$ involving the interaction hand; (2) Apply a mask to conceal the hand, as the target image should not include it, making the hand context mask $M_h$ larger than the detected hand box by a factor of $>1$ (\eg, 1.5 times); (3) Introduce a random context mask $M_r$ with the same scale as $M_h$ to enhance MaskAHand pre-training difficulty, preventing the model from simply predicting the masked region; (4) Apply a random perspective transformation to simulate perspective change between the demonstration video and target image (\eg, egocentric vs. exocentric). 

\begin{figure*}[t]
    \centering
    \includegraphics[width=\linewidth]{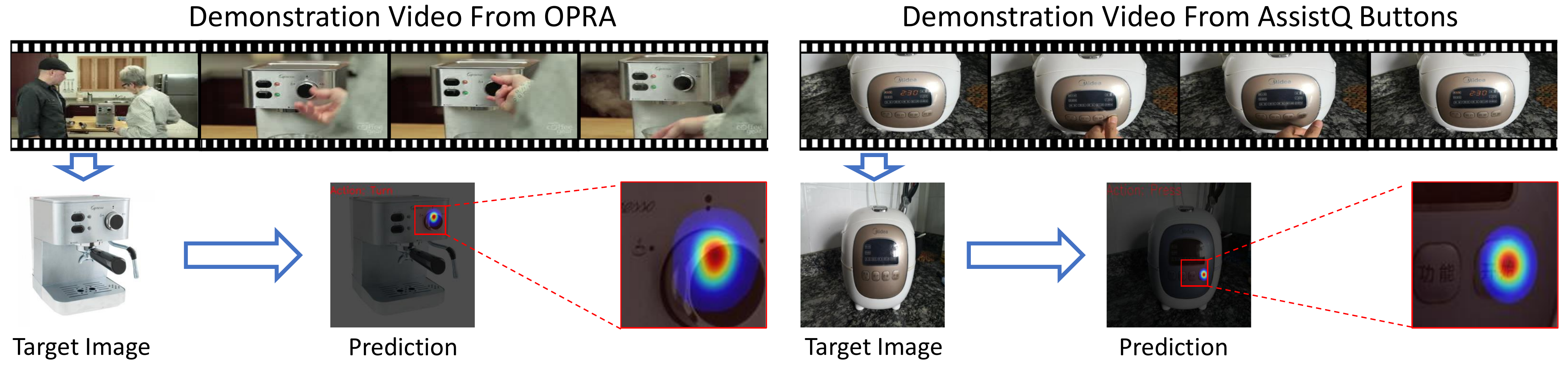}
    \caption{Afformer's video-to-image affordance grounding visualization. MaskAHand's visualization is in the supplementary material. \vspace{-3mm}}
    \label{figure4}
\end{figure*}

\subsection{Video-to-Image Masked Hand Grounding}

Given a masked and perspective-transformed interaction frame as the target image $I$ and a mined interaction clip as the video $V$, the Afformer without action predictor $\boldsymbol{F}'(V, I)$ takes them as input and produces a heatmap $H'$. Unlike the supervised setting, the ground truth for $H'$ is derived from the detected hand box mask, subjected to the same perspective transformation as $I$ (see Figure~\ref{figure3}). During MaskAHand pre-training, the network is tasked with video-to-image masked hand grounding, which requires observing the video to ``match'' the unmasked context in the target image and predict the precise hand box position. Thus, this pre-training shares abilities essential for video-to-image affordance grounding, including context matching between the video and image to ground affordances in the image.

For MaskAHand pre-training, we still utilize the KLD loss, as shown in Equation~\ref{equation11}. The ground-truth hand box masks are gaussian blurred, following video-to-image affordance grounding~\cite{demo2vec}. After training Afformer with MaskAHand, we can directly perform zero-shot evaluation on video-to-image affordance grounding, or fine-tune using supervised data and subsequently conduct evaluation.

\section{Experiments}

\subsection{Experimental settings}

\noindent\textbf{Datasets.} We conduct experiments on three datasets:

\noindent
$\bullet$ \textbf{OPRA}~\cite{demo2vec} consists of YouTube product review videos for appliances (\eg, washing machine, stove). Each data sample consists of a video clip $V$ (\eg, holding a frying pan), paired with a target image $I$ (\eg, ads picture of the frying pan), an action label $A$ (\eg, holding) belonging to a total of $7$ actions, and ten annotated points on the image representing the affordance region (\eg, ten points around the frying pan handle). The ten points are always produced as ground-truth heatmap $H$ by applying a gaussian blur with a kernel size of 3~\cite{demo2vec,hotspot,hagnet}. The dataset comprises roughly $16k$ training samples and $4k$ testing samples, each represented in the form of $(V, I, A, H)$.

\noindent
$\bullet$ \textbf{EPIC-Hotspot}~\cite{hotspot} is made up of EPIC-Kitchens~\cite{epic-kitchens-55}, which contains egocentric videos of kitchen activities. EPIC-Kitchens provides an action label $A$ and annotations for interacted objects in video $V$, but no target image. EPIC-Hotspot chooses one frame in $V$ to be the target image $I$ that corresponds to the object class and appearance. Follow \cite{hotspot} for more information. They crowdsource annotations for ground-truth heatmaps $H$ after $I$ is chosen, yielding $1.8k$ annotated instances across $20$ actions and $31$ objects. The data sample format in EPIC-Hotspot is $(V,I,A,H)$, which is the same as OPRA.

\noindent
$\bullet$ \textbf{AssistQ~\cite{assistq}} is a benchmark to solve user egocentric query~\cite{assistsr} according to instructional videos. It includes fine-grained button regions for a wide range of everyday devices (\eg, microwave), which require precise affordance prediction to distinguish very close buttons. In AssistQ~\cite{assistq}, we consider the instructional video to be the active video and the user view to be the inactive image. Using transcript timestamps, we divide the instructional video into multiple clips $V$ and manually annotate the interacted button on the inactive image $I$. We finally get 650 training samples from 80 videos and 91 testing samples from 20 videos, and each of sample contains active video clip, inactive image, and active-to-inactive button bounding-boxes. The ground-truth heatmap $H$ is generated using gaussian blur for the button center point map. Because the action class in AssistQ is mostly limited to ``press'' or ``turn'', the data sample format is $(V,I,H)$ without action.

We report results on the test sets of these datasets and perform ablation studies on the largest OPRA dataset.

\noindent\textbf{Implementation Details.} We train the Afformer model with a ResNet encoder using a batch size of 16 and 5$k$ iterations, employing the AdamW optimizer~\cite{adamw} and a cosine annealing learning rate scheduler~\cite{cosineanealinglr} with an initial learning rate of $3\times10^{-4}$. As per \cite{demo2vec}, we set the spatial size of both images and videos to 256. The ground-truth heatmap is generated using annotation points (box center for AssistQ) mapped to a Gaussian blur kernel size of $\sqrt{256\times256}/3$, following~\cite{demo2vec,hotspot}. For the Afformer with a ViTDet encoder, we adjust the learning rate to $2\times10^{-4}$ and the spatial size to 1024 to accommodate the pre-trained positional encodings from \cite{vitdet}. All encoders are initialized with COCO~\cite{coco} detection weights~\cite{faster_rcnn,vitdet}, following \cite{demo2vec}. These hyperparameters remain consistent across all datasets, including MaskAHand pre-training.

\noindent\textbf{Evaluation.} We report saliency metrics~\cite{saliency_metrics} as KLD, SIM, and AUC-J. Please refer to \cite{demo2vec,hotspot} for more details.

\begin{table*}[t]
\centering
\footnotesize
\begin{tabular}{c|cc|cc}
\hline
\multirow{2}{*}{Method} & \multicolumn{2}{c|}{\multirow{2}{*}{Variants}} & \multicolumn{2}{c}{OPRA ($256\times256$)} \\
& & & Heatmap KLD $\downarrow$ & Action Top-1 Acc $\uparrow$\\
\hline
\multirow{4}{*}{Demo2Vec~\cite{demo2vec}}  
& \multicolumn{2}{c|}{LSTM} & 3.45 & 20.41 \\
& \multicolumn{2}{c|}{ConvLSTM} & 3.31 & 30.20 \\
& \multicolumn{2}{c|}{TSA + ConvLSTM} & 3.34 & 38.47 \\
& \multicolumn{2}{c|}{Motion + TSA + ConvLSTM} & 2.34 & 40.79 \\
\hline
Naive Baseline (Ours) & \multicolumn{2}{c|}{ResNet-50-Deconv} & 2.20 & 45.66 \\
\hline
\multirow{2}{*}{\shortstack{Afformer (Ours)}}
& \multicolumn{2}{c|}{ViTDet-B-Afformer} & \textbf{1.51} &  \textbf{52.27} \\
& \multicolumn{2}{c|}{ResNet-50-Afformer} & \textbf{1.55} & \textbf{52.14} \\
\hline
\multirow{4}{*}{\shortstack{MaskAHand (Ours)}} & \multirow{2}{*}{Zero-shot} & ResNet-50-Deconv & 2.89 & n/a \\
&& ResNet-50-Afformer & \textbf{2.36} & n/a \\
\cline{2-5}
& \multirow{2}{*}{Fine-tune} & ResNet-50-Deconv & 1.74 & 48.93 \\
&& ResNet-50-Afformer & \textbf{1.48} & \textbf{52.50} \\
\hline
\end{tabular}
\vspace{-1mm}
\caption{Video-to-image affordance grounding performance of our Afformer and MaskAHand models on the OPRA dataset (fine-grained, $256\times256$): Afformer reduces heatmap KLD errors by over 30\%; MaskAHand's zero-shot pre-training results are comparable to \cite{demo2vec} (2.36 \textit{vs.} 2.34); further fine-tuning yields the best performance on OPRA.}
\label{table1}
\end{table*}

\begin{table*}[t]
\centering
\footnotesize
\begin{tabular}{c|c|ccc|ccc}
\hline
\multirow{2}{*}{Method} & \multirow{2}{*}{Method} & \multicolumn{3}{c|}{OPRA ($28\times28$)} & \multicolumn{3}{c}{EPIC ($28\times28$)}  \\
 &  & KLD $\downarrow$ & SIM $\uparrow$ & AUC-J $\uparrow$& KLD $\downarrow$ & SIM $\uparrow$ & AUC-J $\uparrow$\\
\hline
\multirow{6}{*}{\shortstack{Weakly\\Supervised}}  
& EGOGAZE~\cite{egogaze,hotspot} & 2.43 & 0.25 & 0.65 & 2.24 & 0.27 & 0.61 \\
& MLNET~\cite{mlnet,hotspot} & 4.02 & 0.28 & 0.76 & 6.12 & 0.32 & 0.75 \\
& DEEPGAZEII~\cite{deepgaze,hotspot} & 1.90 & 0.30 & 0.72 & 1.35 & 0.39 & 0.75 \\
& SALGAN~\cite{salgan,hotspot} & 2.12 & 0.31 & 0.77 & 1.51 & 0.40 & 0.77 \\
& Hotspot~\cite{hotspot} & 1.42 & 0.36 & 0.81 & 1.26 & 0.40 & 0.79 \\
& HAG-Net (+Hand Box)~\cite{hagnet} & 1.41 & 0.37 & 0.81 & 1.21 & 0.41 & 0.80 \\
\hline
\multirow{2}{*}{\shortstack{Self-supervised \\ Zero-shot}} 
& Center Bias (action agnostic) & 11.13 & 0.21 & 0.63 & 10.66 & 0.22 & 0.63 \\ 
& MaskAHand (action agnostic) & 1.86 & 0.28 & 0.76 & 1.32 & 0.37 & 0.76 \\
\hline
\multirow{3}{*}{Supervised}  
& Img2heatmap~\cite{hotspot} & 1.47 & 0.36 & 0.82 & 1.40 & 0.36 & 0.80 \\
& Demo2Vec~\cite{demo2vec} & 1.20 & 0.48 & 0.85 & n/a & n/a & n/a \\
& \textbf{Afformer~(Ours)} & \textbf{1.05} & \textbf{0.53} & \textbf{0.89} & \textbf{0.97} & \textbf{0.56} & \textbf{0.88} \\
\hline
\end{tabular}
\vspace{-1mm}
\caption{Performance of Afformer and MaskAHand models on OPRA and EPIC-Hotspot datasets (coarse-grained, $28\times28$). MaskAHand can surpass many weakly-supervised methods in KLD. Afformer achieve the best performance among supervised methods. }
\label{table2}
\end{table*}

\begin{table}[t]
\small
\centering
\begin{tabular}{c|ccc}
\hline
Method & KLD $\downarrow$ & SIM $\uparrow$ & Top-1 Acc $\uparrow$ \\
\hline
Afformer & 1.13 & 0.44 & 0.41 \\
+MaskAHand & \textbf{1.01(-12\%)} & \textbf{0.54(+23\%)} & \textbf{0.57 (+28\%)} \\
\hline
\end{tabular}
\vspace{-1mm}
\caption{Results on AssistQ Buttons~\cite{assistq}. ``Acc'' refers to the accuracy of button classification.}
\label{table3}
\end{table}

\subsection{Main Results}

\noindent\textbf{Fine-grained Video-to-image Affordance Grounding.} 
In Table~\ref{table1}, the prior method Demo2Vec~\cite{demo2vec} achieves a KLD error of 2.34 on OPRA, which is comparable to our naive ResNet-50~\cite{resnet} + deconvolution baseline with 2.20 KLD. However, our proposed Afformer significantly reduces KLD error. Utilizing ResNet-50 and ViTDet-B~\cite{vitdet} backbones, Afformer attains 1.55 and 1.51 KLD, respectively, surpassing previous results by over 30\%. We attribute Afformer's success to its better design for fine-grained affordance heatmaps, which also boosts action classification accuracy.

We also assess MaskAHand pre-training as Table~\ref{table1} reveals, surprisingly, its zero-shot results are already comparable to Demo2Vec (2.36 \textit{vs.} 2.34), demonstrating its effectiveness as a proxy task for supervised video-to-image affordance tasks. Furthermore, fine-tuning the MaskAHand pre-trained Afformer on OPRA leads to the lowest KLD error of 1.48, a 37\% improvement over Demo2Vec.

\noindent\textbf{Coarse-grained Affordance Grounding on OPRA, EPIC-Hotspots.} 
We adopt the evaluation protocol from \cite{hotspot} to assess our method's performance on coarse-grained affordance grounding at low resolution ($28 \times 28$). We downsample the standard resolution $256 \times 256$ prediction heatmap to $28 \times 28$ using bilinear interpolation during both training and inference phases. Table~\ref{table2} demonstrates that Afformer, despite not being explicitly designed for lower resolution, outperforms other methods. Moreover, the self-supervised MaskAHand zero-shot results surpass some weakly-supervised approaches.

\noindent\textbf{Video-to-image Grounding on small-scale Data.} 
We train our Afformer and fine-tune MaskAHand models on AssistQ Buttons, which contains only 600+ training samples, posing challenges for data-hungry deep neural networks. As demonstrated in Table~\ref{table3}, MaskAHand pre-training substantially reduces the heatmap KLD error and increases SIM metric, increasing the relative button classification accuracy by 28\%. Thus, MaskAHand self-supervised pre-training is a viable option when dealing with limited video-to-image affordance grounding data.

\begin{table*}
\footnotesize
\centering
\begin{subtable}[t]{0.25\textwidth}
\centering
\begin{tabular}{c|c|c}
\hline
Spatial I & Spatial V & KLD $\downarrow$ \\
\hline
$8^2$ & \multirow{4}{*}{$8^2$} & 1.88\\
$16^2$ &  & 1.73 \\
$32^2$ &  & \textbf{1.65} \\
$64^2$ &  & \textbf{1.65} \\
\hline
\multirow{3}{*}{$32^2$} & $16^2$ & 1.63  \\ 
 & $32^2$ & \textbf{1.62} \\
 & $64^2$ & 1.63 \\ 
 \hline
\end{tabular}
\subcaption{Single pyramid. I: image, V: video.} 
\end{subtable}
%
%
\begin{subtable}[t]{0.35\textwidth}
\centering
\begin{tabular}{c|c|c}
\hline
Spatial I & Spatial V & KLD $\downarrow$\\
\hline
\multirow{2}{*}{$32^2$} & $32^2$ & \textbf{1.62} \\
 & $16^2 \rightarrow 32^2$ & 1.63  \\
\hline
$32^2$ & \multirow{2}{*}{$32^2$} & 1.62 \\
$16^2 \rightarrow 32^2$ & & \textbf{1.60}  \\
\hline
$8^2 \rightarrow 16^2 \rightarrow 32^2$ & \multirow{3}{*}{$32^2$} & 1.59  \\
$16^2 \rightarrow 32^2 \rightarrow 64^2$ &  & \textbf{1.57} \\
$8^2 \rightarrow ... \rightarrow 64^2$ & & \textbf{1.57} \\
\hline
\end{tabular}
\subcaption{Multiple pyramids. I: image, V: video.}
\end{subtable}
%
%
\begin{subtable}[t]{0.35\linewidth}
\centering
\begin{tabular}{c|c|c|cc}
\hline
\multirow{2}{*}{Spatial} & \multirow{2}{*}{Temporal} & \multirow{2}{*}{KLD $\downarrow$} & \multirow{2}{*}{Mem $\downarrow$} \\
& & & \\
\hline
\multirow{6}{*}{\shortstack{$16^2$ \\ $\downarrow$ \\ $32^2$ \\ $\downarrow$ \\ $64^2$}} & \multirow{2}{*}{$T$} & \multirow{2}{*}{1.57} & \multirow{2}{*}{$-0.0$ G} \\
& & & \\
\cline{2-4}
& \multirow{2}{*}{$T \rightarrow \frac{T}{2} \rightarrow \frac{T}{2}$} & \multirow{2}{*}{1.56} & \multirow{2}{*}{$-2.7$ G} \\
& & & \\
\cline{2-4}
& \multirow{2}{*}{$T \rightarrow \frac{T}{2} \rightarrow \frac{T}{4}$} & \multirow{2}{*}{\textbf{1.55}} & \multirow{2}{*}{$-3.8$ G} \\
& & & \\
\hline
\end{tabular}
\subcaption{Temporal pyramids and reduced memory.}
\end{subtable}
%
%
\vspace{-1mm}
\caption{Ablation study results for Afformer on the OPRA dataset at a $256\times256$ scale, comparing default settings (image spatial pyramids: $16^2\rightarrow32^2\rightarrow64^2$, video spatial pyramid: single $32^2$, no temporal downsampling, and one cross/self-attention module in decoding). Enhanced by multi-scale, high-resolution decoding, performance significantly improves (1.57 vs. 1.88), while temporal pyramids further reduce KLD error and decrease GPU memory consumption. \vspace{-2mm}}\label{table4}
\end{table*}

\vspace{-0.2cm}

\begin{table}
\centering
\footnotesize
\begin{subtable}{1.0\linewidth}
\centering
\begin{tabular}{ccc|cc}
\hline
\multirow{2}{*}{\shortstack{\# Hand Mask \\ ($\#M_h<=1$)}} & \multirow{2}{*}{\shortstack{\# Random\\ Mask ($\#M_r$)}} & \multirow{2}{*}{\shortstack{Mask \\ Scale}}  & \multirow{2}{*}{Zero-shot KLD} \\
&&&\\
\hline
0 & 0 & n/a &  4.35 \\
\hline
1 & 0 & 1.0$\times$ & 4.29 \\
1 & 0 & 1.5$\times$ & 2.68  \\
1 & 0 & 2.0$\times$ & 2.54 \\
1 & 0 & 3.0$\times$ & 3.42 \\
\hline
1 & 1 & 1.5$\times$ & \textbf{2.48} \\
1 & 1 & 2.0$\times$ & 2.75 \\
1 & 2 & 2.0$\times$ & 2.98 \\
\hline
\end{tabular}
\subcaption{Ablations on masking ratio and number of masks.}
\end{subtable}
\begin{subtable}{1.0\linewidth}
\centering
\begin{tabular}{c|c|cc}
\hline
Masking & Distortion & Zero-shot KLD & Fine-tune KLD \\
\hline
\multirow{4}{*}{\shortstack{$\#M_h=1$, \\ $\#M_r=1$, \\ 1.5$\times$}} & 0 & 2.48 & 1.53 \\
& 0.25 & 2.40 & 1.50 \\
& 0.5 & \textbf{2.36} & \textbf{1.48} \\
& 1.0 & n/a & n/a \\
\hline
\end{tabular}
\subcaption{Ablations on perspective transformation. ``n/a'': network divergence.}
\end{subtable}
\vspace{-1mm}
\caption{Ablation studies of MaskAHand pre-training on OPRA. \vspace{-2mm}}
\label{table5}
\end{table}

\subsection{Ablation Studies}

\noindent\textbf{Afformer Fine-grained Decoder.} 
We evaluate our fine-grained decoder on the OPRA dataset ($256\times256$ resolution). Table~\ref{table4}(a) reveals that the highest image pyramid resolution ($64^2$) reduces the KLD error; however, for videos, a $32^2$ resolution is more effective, potentially due to weaker semantics in high-resolution pyramids. Table~\ref{table4}(b) indicates that preserving a fixed video pyramids when building low-to-high resolution image pyramids also decreases KLD error. As per Table~\ref{table4}(c), constructing video temporal pyramids results in considerable memory savings and slight performance enhancement. These findings suggest the significance of our fine-grained decoder in heatmap decoding.

\noindent\textbf{Context Masking in MaskAHand.} 
We examine the context masking effects in MaskAHand (Table~\ref{table5}(a)). Without masking ($\#M_h=0$ and $\#M_r=0$), pre-training degenerates into hand saliency detection (Figure~\ref{figure3}), providing no benefit to video-to-image affordance grounding and resulting in a large KLD error 4.35. A similar outcome occurs with hand masking only: when $\#M_h=1$ and $\#M_r=0$, the network can directly predict the masked region for a low training loss, yielding a meaningless result (KLD 4.29).

However, increasing the hand mask region to $1.5\times$ significantly reduces the zero-shot KLD error to $2.68$, indicating that the network learns a useful representation for video-to-image affordance grounding. Extending the mask scale to $3.0\times$ causes performance degradation, likely due to the overly large masked region creating challenging context matching between video and image. An additional random mask offers a similar effect as enlarging the hand mask but with more diversity, preventing the network from merely predicting a region within the hand mask. Therefore, we use $\#M_h=1$, $\#M_r=1$, and $1.5\times$ hand box masking as the default MaskAHand pre-training setting.
 
\noindent\textbf{Perspective Transformation in MaskAHand.} 
Our context masking strategy is to enhance the grounding ability across the video-image context differences. However, it is also crucial to consider perspective transformation, such as enabling simulation of egocentric and exocentric views. As shown in Table~\ref{table5}(b), perspective transformation can lead to performance improvements when the distortion ratio is in a reasonable range.

\section{Conclusion}
In this paper, we introduce the Affordance Transformer (Afformer), a simple and effective model for video-to-image affordance grounding, utilizing multi-scale decoding to generate fine-grained affordance heatmaps. We also propose a pre-training technique, Masked Affordance Hand (MaskAHand), that employs a proxy task of masked hand interaction grounding, facilitating data collection while benefiting video-to-image affordance grounding. Our extensive experiments show Afformer significantly outperforms previous methods, and MaskAHand pre-training impressively improves performance on small datasets.

\vspace{2mm}
\noindent\textbf{\large{Acknowledgment}} \quad
This project is supported by the National Research Foundation, Singapore under its NRFF Award NRF-NRFF13-2021-0008, and Mike Zheng Shou’s Start-Up Grant from NUS. The computational work for this article was partially performed on resources of the National Supercomputing Centre, Singapore.

{\small
\bibliographystyle{ieee_fullname}
\bibliography{ai}
}

\clearpage

\noindent\large{\textbf{Supplementary Material}}
\vspace{1mm}

We provide supplementary material to complement the main paper. The contents include:

$\bullet$ Section~\ref{sectionb}: We supplement the implementation details of main paper. 

$\bullet$ Section~\ref{sectionc}: We present the additional ablation studies for Afformer and MaskAHand.

$\bullet$ Section~\ref{sectiond}: We show MaskAHand generated samples on OPRA~\cite{demo2vec}, EPIC-Hotspot~\cite{hotspot}, and AssistQ~\cite{assistq} Buttons.

$\bullet$ Section~\ref{sectione}: We present MaskAHand zero-shot visualization results on OPRA test set.

\begin{table}[h]
\footnotesize
\centering
\begin{subtable}{\linewidth}
\centering
\begin{tabular}{l|l}
\hline
    Config & Value \\
    \hline
    Optimizer & AdamW, weight decay $0.05$ \\
    Learning rate & $3\times10^{-4}$, cosine decay \\
    Training batch size \& epochs & 16 batch size in 7 epochs \\ 
    Automatic mixed precision & 16-bit \\
    Backbone learning rate factor & 0.1 \\
    Backbone Initialization & COCO Detection~\cite{faster_rcnn,coco} \\
    Attention Heads \& Channels & 4 \& 256 \\
    Maximum Video Frames & 64 by uniform sampling\\
    \hline
\end{tabular}
\subcaption{Setting for supervised training of Afformer on OPRA~\cite{demo2vec}, EPIC-Hotspots~\cite{hotspot}, and AssistQ~\cite{assistq} Buttons. Note that EPIC-Hotspots and AssistQ Buttons are only experimented with R50-FPN backbone. \vspace{2mm}}
\end{subtable}
\begin{subtable}{\linewidth}
\centering
\begin{tabular}{l|l}
\hline
    Config & Value \\
    \hline
    Backbone & R50-FPN~\cite{resnet,fpn} \\
    Batch size & 32 \\
    Mined video frames & 32 by sequential sampling\\
    Stride between mined videos & 16 frames \\
    Interaction score threshold & 0.99 \\
    \hline
\end{tabular}
\subcaption{Setting for MaskAHand pre-training on OPRA~\cite{demo2vec}, EPIC-Hotspots~\cite{hotspot}, and AssistQ~\cite{assistq} Buttons. Other settings follow (a).}
\end{subtable}
\caption{Implementation details of Afformer and MaskAHand.}\label{tablea}
\end{table}

\begin{table}[h]
\footnotesize
\centering
\begin{subtable}{\linewidth}
\centering
\begin{tabular}{cc|c}
\hline
Module & I,V Shared & KLD $\downarrow$ \\
\hline
\multirow{2}{*}{Backbone} & $\times$ & 1.80 \\
 & \checkmark & \textbf{1.57} \\
\hline
\multirow{2}{*}{Input Proj} & $\times$ & 1.65 \\
& \checkmark & \textbf{1.57} \\
\hline
\end{tabular}
\subcaption{Sharing parameters for image and video encoding has better results. \vspace{2mm}}
\end{subtable}
\begin{subtable}{\linewidth}
\centering
\begin{tabular}{cc|c}
\hline
Module & Pyramid Shared & KLD $\downarrow$ \\
\hline
\multirow{2}{*}{Input Proj} & $\times$ & \textbf{1.57} \\
 & \checkmark & 1.60 \\
\hline
\multirow{2}{*}{Decoder} & $\times$ & 1.61 \\
& \checkmark & \textbf{1.57} \\
\hline
\end{tabular}
\subcaption{Different pyramids need different input projections, but the decoder can be shared.}
\end{subtable}
\caption{Additional ablation studies for Afformer on OPRA.}\label{tableb}
\end{table}

\begin{table}[h]
\centering
\footnotesize
\begin{tabular}{c|cc}
Mask & Zero-shot KLD $\downarrow$ & Fine-tune KLD $\downarrow$ \\
\hline
Zero ($0$) & 2.65 & 1.50 \\
Random ($0\sim255$) & \textbf{2.36} & \textbf{1.48} \\
\end{tabular}
\caption{Experiments on masked fill value of MaskAHand pre-training on OPRA.}\label{tablec}
\end{table}

\section{Implementation Details}\label{sectionb}

Due to limited pages, some implementation details are not presented in the main paper. We complement them in Table~\ref{tablea}. The initialization for Afformer's backbone follows Demo2Vec\cite{demo2vec}, which also initializes the backbone network by COCO detection~\cite{coco} weights. We use \texttt{PyTorch}~\cite{pytorch} and \texttt{torchvision} to implement our model. 

\section{Additional Ablation Studies}\label{sectionc}

Table~\ref{tableb} shows the additional ablation studies for Afformer, which suggests that the backbone and input projector should be shared for video and target image. Furthermore, we observe that the input projector should be different for different pyramid resolution levels, but the transformer decoder can be shared. The sharing of the backbone and decoder makes our Afformer parameter efficient. 

We also investigate the masked filled value of MaskAHand pre-training, as shown in Table~\ref{tablec}. In the masked region of the target image, filling random value (\ie random noise mask) is much better than zero value (\ie black mask). We hypothesize that the fixed zero filling makes model overfitting easier.

\section{Training Samples from MaskAHand}\label{sectiond}

\begin{figure*}[t]
    \centering
    \begin{subfigure}{\linewidth}
        \centering
        \includegraphics[width=\textwidth]{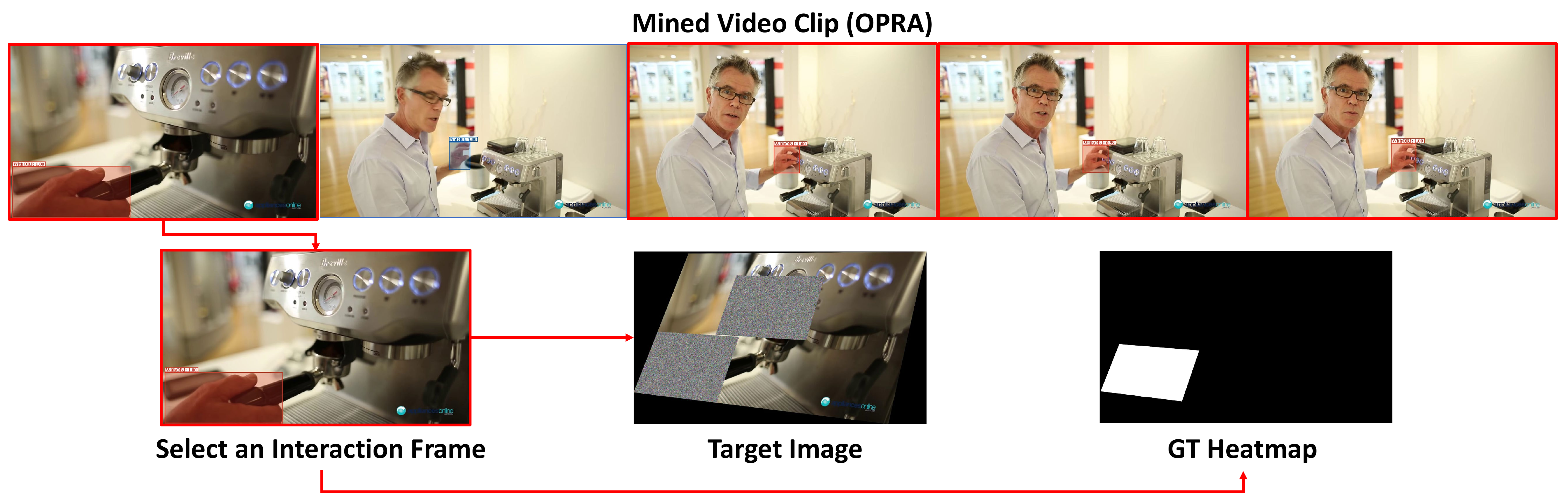}
        \caption{A training sample case generated by MaskAHand, on OPRA training set's raw videos.}
    \end{subfigure}
    \begin{subfigure}{\linewidth}
        \centering
        \includegraphics[width=\textwidth]{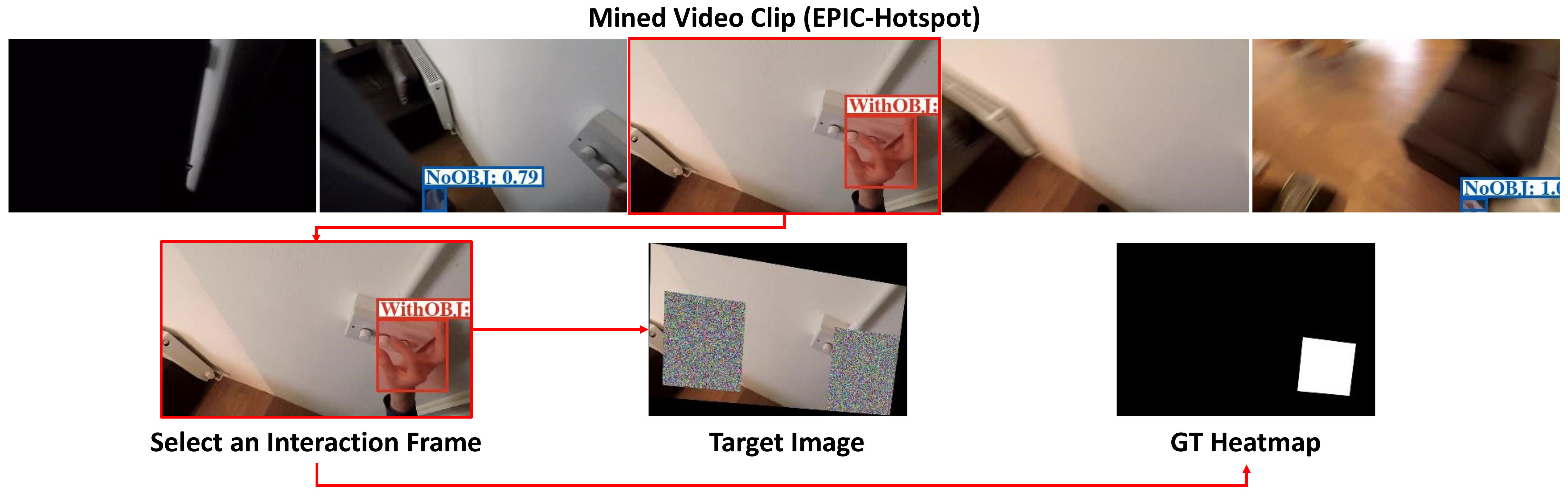}
        \caption{A training sample case generated by MaskAHand, on EPIC-Hotspots training set's raw videos (\ie EPIC-KITCHEN frames~\cite{epic-kitchens-55}).}
    \end{subfigure}
    \begin{subfigure}{\linewidth}
        \centering
        \includegraphics[width=\textwidth]{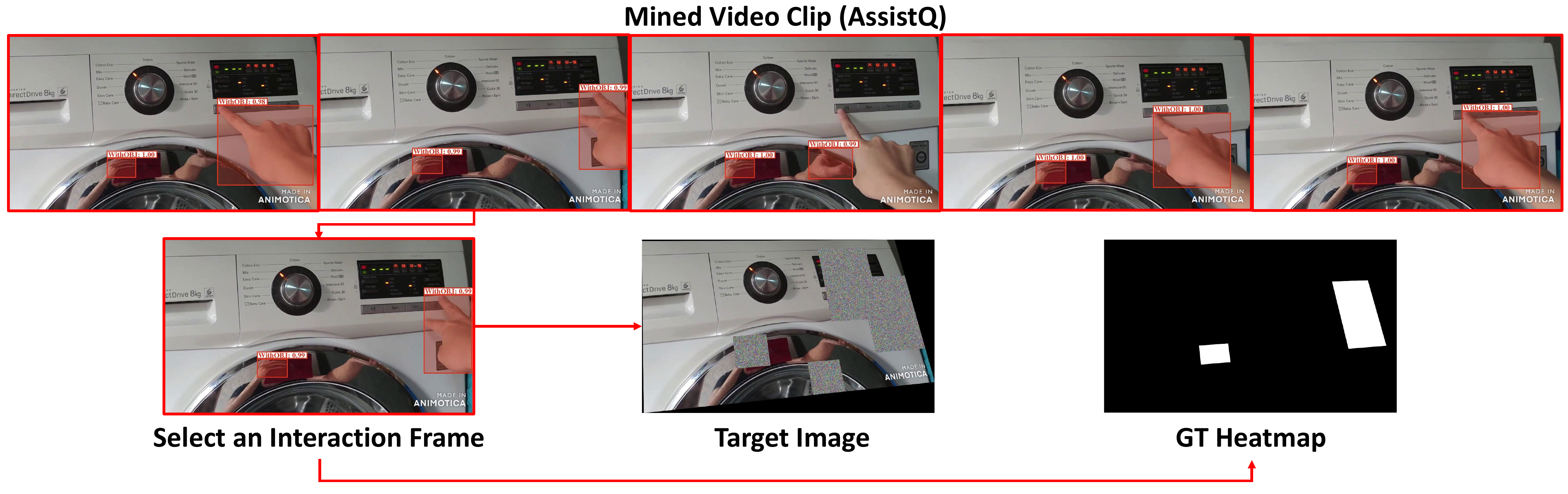}
        \caption{A training sample case generated by MaskAHand, on AssistQ training set's raw videos.}
    \end{subfigure}
    \caption{MaskAHand generated training sample cases on OPRA~\cite{demo2vec}, EPIC-Hotspots~\cite{hotspot}, and AssistQ~\cite{assistq}. The hand interaction detection is made by our trained hand interaction R-CNN, mentioned in Section 4.2. ``WithObj'' means the hand is interacting with objects, whereas ``NoObj'' means not interacting. The mined video clip contains 32 frames (with 5 FPS). If there are multiple interaction frames detected (\eg, (a) and (c)), the target image generation will be randomly picked from these frames.}\label{figureb}
\end{figure*}

Figure~\ref{figureb} shows MaskAHand generated training samples on different datasets. Each training sample include a video clip, a target image, and a ground-truth heatmap.

\section{MaskAHand Zero-shot Visualization}\label{sectione}

\begin{figure*}[t]
    \centering
    \includegraphics[width=\textwidth]{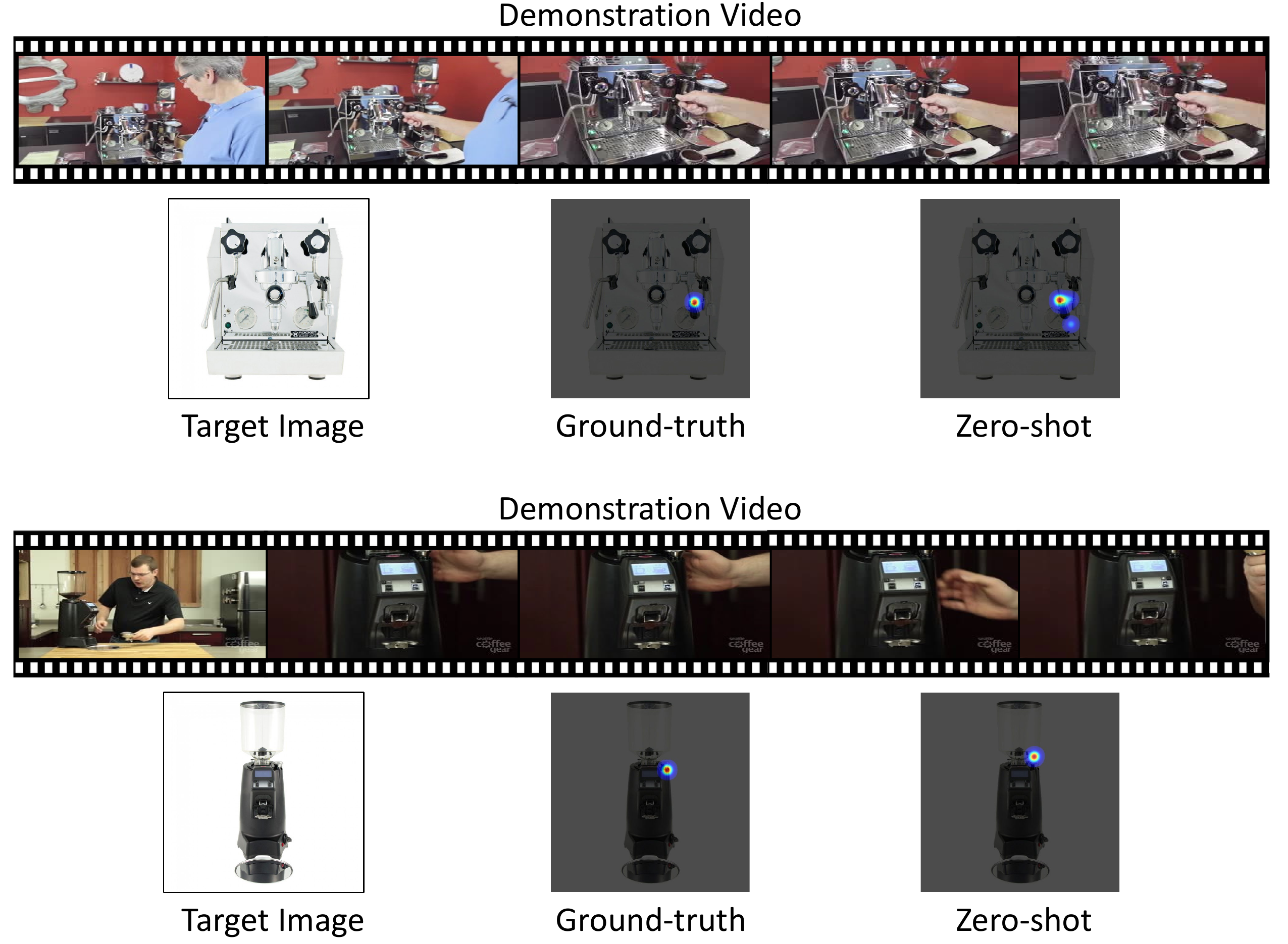}
    \caption{MaskAHand zero-shot results on OPRA test set. For visualization, we select the top 100 points on the ground-truth and zero-shot prediction heatmap. It can be seen that the hotspot on the zero-shot heatmap is close to that on the ground-truth heatmap.}\label{figurec}
\end{figure*}

Figure~\ref{figurec} visualizes MaskAHand zero-shot results, demonstrating that the representation learned by MaskAHand can support video-to-image affordance grounding.

\end{document}